# *UPAQ*: A Framework for Real-Time and Energy-Efficient 3D Object Detection in Autonomous Vehicles

Abhishek Balasubramaniam, Febin P Sunny, Sudeep Pasricha
Department of Electrical and Computer Engineering
Colorado State University, Fort Collins, CO, USA
{abhishek.balasubramaniam, febin.sunny, sudeep}@colostate.edu

*Abstract*— To enhance perception in autonomous vehicles (AVs), recent efforts are concentrating on 3D object detectors, which deliver more comprehensive predictions than traditional 2D object detectors, at the cost of increased memory footprint and computational resource usage. We present a novel framework called *UPAQ*, which leverages semi-structured pattern pruning and quantization to improve the efficiency of LiDAR point-cloud and camera-based 3D object detectors on resource-constrained embedded AV platforms. Experimental results on the Jetson Orin Nano embedded platform indicate that *UPAQ* achieves up to 5.62× and 5.13× model compression rates, up to 1.97× and 1.86× boost in inference speed, and up to 2.07× and 1.87× reduction in energy consumption compared to state-of-the-art model compression frameworks, on the Pointpillar and SMOKE models respectively.

*Keywords—3D object detection, pruning, quantization, Jetson Orin Nano, model compression, pointpillar, smoke, machine learning.*

## I. INTRODUCTION

Autonomous vehicles (AVs) are essential for enhancing transportation efficiency and significantly reducing crash-related injuries and fatalities [1]. Emerging AVs rely on advanced perception systems—integrating object classification, 3D positioning, and object detection using sensors like LiDARs and cameras. In particular, object detectors (ODs) play a vital role in emerging AVs, being responsible for perceiving an AV's surrounding environment based on captured sensor data and serving as the foundation for the subsequent decision-making process. Given their relevance in ensuring safety and executing safety-critical tasks, ODs must deliver high accuracy and achieve real-time inferences within tens of milliseconds [2]. While there have been many advances in OD accuracy in recent years, these improvements have increased memory footprint and computational overheads on embedded AV platforms [3]. These embedded platforms in AVs also need to process data from multiple on-board systems, V2X communication, and infotainment, which escalates computational demands and reduces power headroom [4].

Modern AVs are increasingly relying on rich 3D data, referred to as pointcloud data, from their sensors and utilizing 3D ODs which can provide depth, size, and location information that 2D ODs cannot. Recent advances have led to complex 3D ODs leveraging sophisticated algorithms for improved accuracy [5]. However, this complexity increases memory usage and computational efficiency compared to the best 2D ODs, reducing real-time performance. To tackle this challenge, advanced model compression techniques have emerged, including pruning, quantization, knowledge distillation, and low-rank factorization [6]. Among these, pruning stands out for its extensive application in machine learning, increasing parameter sparsity and significantly reducing computational costs [2]. However, conventional pruning methods fall short when considering essential performance metrics such as latency, memory usage, and energy consumption during model execution [7]. This highlights the need for more effective solutions to enhance 3D OD performance.

In this paper, we present the *UPAQ* framework, designed for efficient 3D OD compression, via a two-tier compression technique, which leverages kernel pruning and kernel quantization. Central to our approach is a novel model optimization strategy which optimizes the model sparsity and bitwidths while preserving overall accuracy. The novel contributions of our *UPAQ* framework are as follows:

- A model compression approach that retains key feature maps and performs high accuracy real-time 3D object detection with both 3D pointcloud LiDAR data and 3D camera data;
- A kernel transformation approach for quantizing and pruning 1x1 kernels for better generalization of 3D pointcloud features;
- A hybrid approach for mixed precision quantization with semi-structured pruning for improved accuracy retention;
- A model of on-device efficiency of the compressed model to select the best fit quantized kernels for the model;
- A detailed comparison with state-of-the-art OD pruning and quantization methods, demonstrating the effectiveness of our framework in terms of mean average precision (mAP), latency, energy efficiency, and sparsity.

## II. RELATED WORK

Before the emergence of 3D ODs, object detection was largely handled through 2D ODs. 2D ODs function by generating bounding boxes on a two-dimensional plane defined by four coordinates: [$x_{min}$, $y_{min}$, $x_{max}$, $y_{max}$]. There are two main categories of 2D ODs. *Two-stage detectors* operate in two steps, first generating region proposals and then classifying those proposals. Examples include Fast R-CNN [8] and Faster R-CNN [9]. These models are extremely resource intensive [11] and have low throughput due to the separate stages involved in object recognition. *Single-stage detectors* are designed for low latency, functioning with a single feed-forward network. Examples include RetinaNet [10] and YOLOX [12]. However, all of these (and other) 2D OD models do not possess the ability to perceive depth and cannot utilize the 3D data that a modern sensor suite in AVs can provide.

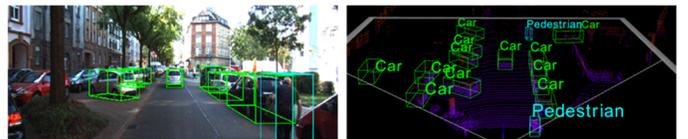

**Fig. 1.** 3D object detection example on SMOKE (left) and Pointpillars (right).

3D ODs address the limitations of their 2D counterparts by offering bounding boxes with nine degrees of freedom, incorporating three positional, three dimensional, and three rotational parameters. 3D ODs typically rely on LiDAR pointclouds, RGB cameras, or a combination of these to enhance their ability to interpret their surroundings. *Pointcloud-based 3D OD models* utilize LiDAR data in a 3D coordinate system, offering high detection accuracy, especially for hard-to-detect objects. These models can process 3D LiDAR data directly (e.g., PointNet [13]) or convert it to 2D (e.g., PointPillars [14]), though the latter may lose features for higher throughput. *RGB camera-based 3D OD models* use images for semantic information but lack depth information. So, they typically employ a two-step process of 2D detection followed by 3D bounding box conversion. Examples include Monoflex [15] and SMOKE [16], which are computationally simpler than pointcloud-based models but less accurate, as shown in Fig. 1, where SMOKE is unable to detect many objects in the foreground and background. However, even though pointcloud-based 3D OD models offer higher precision outputs, they have higher computational complexity and a large memory footprint. A few prior efforts have devised lightweight neural networks for OD, such as SECOND [17] which uses a sparse voxel grid, Focals Conv [18] which uses sparse convolutional layers to speed up inference of pointclouds by focusing on regions with significant data, and VSC [19] which uses a virtual convolution to efficiently process sparse pointcloud data. However, there are still significant challenges to overcome, including the need to more aggressively optimize memory usage and computational efficiency while ensuring high accuracy in diverse and complex environments. The trade-offs between model size and performance (see Table 1) remain critical, particularly as applications demand real-time processing capabilities. Therefore, ongoing research must not only enhance the architectures of sparse networks but also develop advanced techniques for compression, generalization, and robust multimodal integration to further advance the field of real-time 3D OD.

Table 1: Comparison of 3D OD model sizes vs execution time

| Models | Number of parameters (Millions) | Execution time (ms) |
|---|---|---|
| PointPillar [14] | 4.8 | 6.85 |
| SMOKE [16] | 19.51 | 30.65 |
| SECOND [17] | 5.3 | 9.83 |
| Focals Conv [18] | 13.70 | 26.5 |
| VSC [19] | 24.5 | 40.56 |

Prior works such as Ps and Qs (PQ) [20], Clip-Q [21], LIDAR-PTQ [22], and R-TOSS [23] have proposed utilizing quantization and/or unstructured and structured model pruning techniques to compress models and reduce their overheads. The authors in [20] employ Quantization-Aware Training (QAT) with unstructured pruning, but their approach has long training times and reduced model stability. The approach in [21] also combines QAT with unstructured pruning. However, it lacks efficient convergence techniques for balancing accuracy and latency, utilizing a partitioning approach that focuses on only parts of the model without considering overall performance. The approach in [22] uses Post-Training Quantization (PTQ) to convert 32-bit floating-point weights to 8-bit integers, which can enhance the model's inference throughput, but this also reduces accuracy. A recent work [23] employed semi-structured pruning with predefined kernel patterns, known as entry patterns (EPs) to trade-off between structured and unstructured pruning. While promising, the approach lacks the specificity needed to preserve critical features, and the proposed pruning pattern mapping approach leads to suboptimal quantizable weight patterns that compromise inference accuracy. Furthermore, the L2-norm used for selecting the best kernel mask does not adequately account for quantization noise, which further adversely affects performance. In summary, to improve model performance while minimizing latency and maintaining accuracy, it is essential to integrate both pruning and quantization techniques effectively. Integrating advanced pruning with quantization methods that account for quantization noise, such as adaptive kernel mask selection, can improve model efficiency and preserve feature extraction accuracy.

Our proposed *UPAQ* framework implements such an integrated quantization and pruning approach. Specifically, *UPAQ* supports a semi-structured pattern-based pruning method alongside mixed precision quantization. This strategy increases model sparsity more effectively than unstructured approaches, preserving critical weights in the kernels for improved accuracy. Additionally, modern point-cloud-based 3D ODs employ pointcloud-to-pseudo-image conversion techniques, such as Pillar Feature Networks [13] [14], to decrease computational complexity during detection. These networks utilize 1×1 convolutional layers to transform and normalize 3D data into a 2D plane, necessitating high precision. Traditional methods that fix the values of these 1×1 convolutional layers during quantization can diminish model accuracy in the earlier layers, leading to overall performance degradation. Our approach addresses this issue by dynamically adjusting the 1×1 kernel weights, thus preserving accuracy during the detection phase while simultaneously creating a sparsity-aware quantized model that effectively reduces the overall footprint.

## III. Model Compression Background

### A. Model Pruning

Pruning is a widely used model compression strategy that promotes sparsity in neural networks through various regularization techniques. This method typically reduces both memory footprint and computational costs while maintaining accuracy. The computational cost of a model can be expressed as:

$$Computational\ cost\ (C) = (L_n \times K_n \times W_n) \quad (1)$$

where $L_n$ is the number of convolutional layers, $K_n$ denotes the number of kernels in a layer, and $W_n$ represents the number of non-zero weights. As sparsity in the model increases, the computational cost ($C$) decreases. Recent advances in embedded computing platforms have introduced hardware support for compressing weight matrices during inference, allowing for the omission of zero weights, thus reducing model latency when pruning is employed [23].

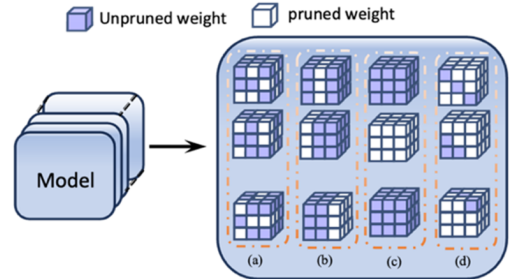

**Fig. 2.** Illustration of different pruning methods (a) unstructured pruning, (b) channel pruning, (c) filter pruning and (d) semi-structured pattern pruning.

Pruning methodologies can be classified into three primary categories: *1) Unstructured Pruning*: This technique selectively prunes weights to minimize model loss while maintaining accuracy (Fig. 2(a)). Algorithms in this category include weight magnitude pruning [24], gradient magnitude pruning [25], and second-order derivative pruning [26]. However, these methods can disrupt thread-level parallelism due to load imbalances from varying sparsity levels and may impair memory performance by altering data access patterns, reducing caching efficiency on GPUs, CPUs, and TPUs [26], [27]; *2) Structured Pruning*: This method systematically removes entire channels (Fig. 2(b)) or filters (Fig. 2(c)) to enhance model sparsity. By creating a uniform weight matrix, filter and channel pruning can significantly reduce multiply-accumulate (MAC) operations compared to unstructured pruning [27]. Structured pruning can be integrated with acceleration frameworks like TensorRT [29], which can use the uniform pruned structures to optimize hardware acceleration across diverse platforms [28]. However, this approach often decreases model accuracy, as essential weights may be pruned alongside redundant ones; *3) Pattern-based Semi-Structured Pruning*: This approach combines structured and unstructured pruning aspects (Fig. 2(d)). It uses kernel masks to selectively retain specific weights, inducing partial sparsity within a kernel. The efficacy of pruned kernels can be evaluated using metrics like the L2-norm. Since kernel patterns are limited to a fixed number of pruned weights, they generally achieve lower sparsity than fully structured or unstructured methods. Connectivity pruning can address this limitation by fully pruning specific kernels [23], [30]. However, pattern pruning often targets kernels of size 3×3 and larger, providing more candidate weights for pruning. Connectivity pruning can end up reducing model accuracy by removing critical weights from kernels. Nonetheless, the semi-structured nature of pattern pruning enables effective hardware parallelism, reducing inference times [23].

### B. Model Quantization

Quantization is a model optimization technique that reduces memory footprint and computational costs by converting weights (and optionally activations) from higher floating-point precision to lower precision. With advancements in hardware and software, many platforms now support precision levels as low as 1-bit integers.

Approaches for model quantization can be divided into two types based on when quantization occurs during model development: *1) Quantization Aware Training (QAT)* involves adding quantization and de-quantization nodes to a fully trained model and retraining it for a set number of epochs. The model calculates the scale factor and simulates quantization loss, integrating it into the overall loss function through fine-tuning, which enhances robustness for subsequent Post-Training Quantization [31]; *2) Post-Training Quantization (PTQ)* quantizes a fully converged model using various algorithms and precision levels. Quantization can also be classified into two categories based on how parameters are changed [32]: *1) Integer quantization*: Here a calibration dataset is used to convert 32-bit floating-point parameters to fixed-point integers (e.g., 8-bit) by calculating the minimum and maximum values of model parameters using the calibration dataset; *2) Floating point quantization*: This method reduces precision from 32-bit to lower bitwidth (e.g., 16-bit) floating-point tensors, prioritizing model accuracy compared to integer quantization. In most approaches, a common practice is to quantize all layers in a model to the same bit-width. However, for many models there is a distinct difference in sensitivity to quantization from layer to layer. *Mixed precision quantization* addresses this issue by keeping more sensitive layers at higher precision while maintaining the rest of the model in lower bits, effectively improving the performance-efficiency trade-off [32].

## IV. UPAQ FRAMEWORK

In this section, we describe our novel 3D OD model compression framework and provide a detailed algorithmic description of our kernel pruning and quantization techniques. Our compression framework, *UPAQ* (Fig. 3), combines a semi-structured pattern pruning scheme with mixed precision quantization, while incorporating various optimizations to reduce computational costs for 3D OD models. The *UPAQ* framework has three stages: pre-processing, pattern generation, and compression. These are discussed next.

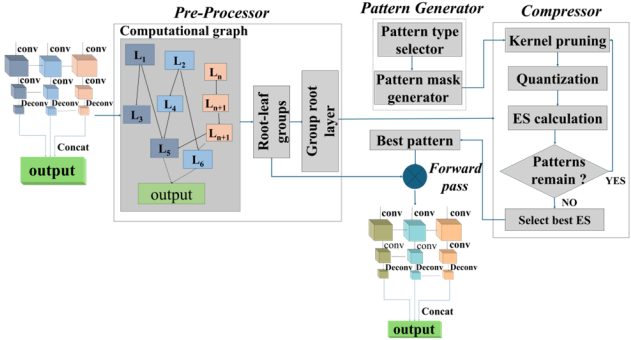

**Fig. 3.** An overview of the proposed *UPAQ* optimization framework.

### A. Pre-processing Stage

In this first stage, we begin by calculating the computational graph of the pretrained model $M$ and utilize the depth-first search (DFS) algorithm to determine the computation paths or connected layers. A root layer may be shared among multiple leaf layers, which we leverage to significantly lower the computational cost associated with optimizing the model. *UPAQ* applies key optimizations to the root layers, so that they are reflected in the leaf layers through forward passes, rather than managing individual layers. The preprocessing stage identifies the root layer-leaf layer subsets within the computation graph so that the optimization steps can be performed with lowered computation cost in the later compression stage (Subsection C).

Algorithm 1 outlines the pseudocode for the preprocessing stage. We start by computing the computational graph $G$ of the model through backpropagation (**line 1**). This graph $G$ is used to execute DFS via the *find_root* function by traversing model layers $l$ and identifying the root for each current layer. A layer can be classified as the root layer when it does not have any other layer designated as its root ($l_r$), indicating that it becomes its own root (**line 4**). This root layer is incorporated into the list of groups initialized as *groups_int* (**line 2**). If a layer is identified as belonging to an existing group, it adopts the corresponding root layer ($l_r$) and is added to that group (**lines 5-6**). Each root layer ($l_r$) can have multiple associated layers, but each of these layers can only be linked to one root layer. This process continues until all layers are categorized into a group. The layers within each group share kernel properties due to their interconnected channels, allowing them to adhere to the same optimization pattern.

**ALGORITHM 1: PREPROCESSING**

*Inputs:* Baseline pretrained model (M)
*Output:* A list of root-leaf layer groups (group)
1  $G \leftarrow$ compute_graph(M)
2  *groups_int* $\leftarrow$ {}
3  **for** *l* in *M*:
4      $l_r \leftarrow$ find_root (G, l)
5      **if** : *[l_r]* in *groups_int* **then:** *groups_int [l_r]* $\leftarrow$ [M [l]];
6      **else:** *groups_int [ l_r ]* $\leftarrow$ 0; *groups_int [ l_r ]* $\leftarrow$ [M [l]];
7  **end**

### B. Pattern Generation Stage

The pattern generator algorithm generates a random pattern of non-zero elements within a $k \times k$ kernel, utilizing one of four potential arrangements: *main-diagonal, anti-diagonal, row,* or *column*. This combined with our compression stage algorithm (Subsection C) ensures that we can obtain the best possible compression for our optimized model, compared to relying on a dictionary of patterns.

**ALGORITHM 2: PATTERN GENERATOR**

*Inputs:* number of non-zero weights (n), dimension of kernel (d)
*Outputs:* Kernel pattern
1  **pattern_type** $\leftarrow$ random choice from ['main_diagonal', 'anti_diagonal', 'row', 'column']
2  **positions** $\leftarrow$ []
3  **if** *pattern_type* = 'main_diagonal' **then**
4      *positions* $\leftarrow$ [(i, i) **for** i **in** (0, min(n, d)]]
5  **else if** *pattern_type* = 'anti_diagonal' **then**
6      *positions* $\leftarrow$ [(i, d -i -1) **for** i **in** (0, min(n, d)]]
7  **else if** *pattern_type* == 'row' **then**
8      *row* $\leftarrow$ random choice from (0, d]
9      *start_col* $\leftarrow$ random choice from (0, d-n]
10     *positions* $\leftarrow$ [(row, start_col + i) **for** i **in** (0, n]]
11 **else if** *pattern_type* == 'column' **then**
12     *col* $\leftarrow$ random choice from (0, d]
13     *start_row* $\leftarrow$ random choice from (0, d-n]
14     *positions* $\leftarrow$ [(start_row + i, col) **for** i **in** (0, n]]
15 **end**

Algorithm 2 outlines the pattern generation approach. First, we randomly select one of the patterns from the predefined list (**line 1**). This selection will dictate how we place the non-zero elements in the kernel. We then initialize *positions* an empty array which will store the coordinates of the non-zero weights in a kernel (**line 2**). If the chosen pattern is the *main_diagonal*, we calculate the position of the non-zero elements as *(i, i)* for the range of 0 to *min(n-d)-1* where *n* and *d* are the number of non-zero weights and dimension of the kernel (**line 3-4**). When the selected pattern is the *anti-diagonal*, we place the non-zero elements at *(i, d-i-1)* for the range of 0 to *min(n-d)-1* (**line 5-6**). In the case where a *row* pattern is selected, we choose a random row of range *(0, d-n]*. We then choose a starting column (*start_col*) along which the non-zero weights are placed which is then used to select the position of the elements using *(row, start_col + i)* for the range of *(0, n]* (**line 7-10**). For instance, if we randomly select column 0 as *start_col*, and if three non-zero elements are needed, we fill all three columns in the selected row, resulting in positions *(row, 0)*, *(row, 1)*, and *(row, 2)*. Finally, if we select a *column* pattern we choose a random column of range *(0, d]*. We then choose a starting row (*start_row*) along which the non-zero weights are placed which is then used to select the position of the elements using *(start_row + i, col)* for the range of *(0, n]* (**line 11-14**). For example, if three non-zero elements are required, we fill all three rows in the chosen column, resulting in positions (*start_row, col*), (*start_row + 1, col*), and (*start_row + 2, col*). The output of this algorithm is a set of coordinates representing locations where the non-zero elements will be placed in the kernel. This pattern selection ensures that the kernel mask is varied and avoids symmetry along diagonal/row/column depending on the *pattern_type* selected.

### C. Compression Stage

In this final stage, we utilize a combination of pruning and quantization to minimize the model footprint. The process includes: *1)* executing semi-structured pattern pruning; *2)* quantizing the weights of the pruned model; and *3)* performance modeling of the compressed model operation on-device, to evaluate the impact on on-device model performance. This is an optimization stage which ensures the best possible model compression while optimizing model performance in terms of accuracy, latency, and energy consumption.

**ALGORITHM 3: COMPRESSION STAGE FRAMEWORK**

    **Inputs:** Pretrained model (*M*), group_int
    **Outputs:** compressed model ($M_C$)
1  $M_C \leftarrow$ deepcopy(M)
2  **for** *l* in $M_c$:
3    **if** *l* in group_init[$l_r$].keys() **then**
4      *weights = weights in l*
5      **for** $K_W$ *in weights:*
6        *shape* $\leftarrow K_W$.*shape*
7        **if** *shape[-1] > 1* **then**
8          *Call Algorithm 4 to perform k×k kernel compression*
9          *Apply the same compression pattern to all kernels in leafnode*
10        **else**:
11          *Call Algorithm 5 to perform 1×1 kernel compression*
12          *Apply the same compression pattern to all kernels in leafnode*

Algorithm 3 outlines our overall framework for compressing a pretrained model *M* by pruning and quantizing its kernels to produce a compressed model $M_C$. The process begins by creating $M_C$ as a deep copy of *M* (**line 1**). Using a deep copy [34] is critical here because it allows us to modify the structure and weights of $M_C$ independently of *M*, preserving the original model for comparisons. This separation is essential in model compression tasks to evaluate the compressed model effectiveness without altering the baseline model structure and performance. We then iterate through each layer *l* in $M_C$ (**line 2**), checking if the layer is part of the root layer of the *group_init* (**line 3**). For each layer that qualifies, it retrieves the weights (**line 4**) and examines each kernel $K_W$ (**line 5**). We then check the dimension of the kernel. If the kernel shape is not *1×1* (**line 7**), it applies Algorithm 4 to perform *k×k* kernel compression (**line 8**) and replicates the same compression pattern for all kernels in the corresponding leaf node (**line 9**). Conversely, if the kernel is *1×1* (**line 10**), it uses Algorithm 5 for compression (**line 11**) and applies the compression pattern to all leaf node kernels (**line 12**). This method ensures an efficient compression strategy tailored to the structure of the model's kernels.

*1) Kernel Compression*

Our framework above (Algorithm 3) requires efficient compression of *k×k* and *1×1* kernels, the approach for which is discussed next.

The *k×k* kernel compression algorithm (Algorithm 4) performs kernel-wise compression through pruning and quantization. It makes use of a mixed precision quantizer (Algorithm 6) and calculates an efficiency score ($E_S$ from eq. (2); discussed later) after applying the compressed kernel back to the model. In Algorithm 4, we set several variables: *best_Es, bestfit_kernel,* and *temp_kernel* (**lines 1-3**). Patterns are generated using *pattern_generator* (Algorithm 2; **line 4**) to induce sparsity in the kernels. Next, we iterate through the kernel weights ($K_W$) of the root layer and utilize the positions (*rows* and *cols*) of the non-zero kernels identified from the generated patterns (**lines 5-7**). These row and column positions are used to establish the locations of the non-zero kernels, which are then processed by *mp_quantizer* (Algorithm 6) for quantization with different quantization bitwidths (*q*). The compressed kernel is applied back to the kernel weights (**lines 8-10**). After this, the $M_c$ is used to calculate the $E_S$ (from eq. (2); discussed later), iterating through all generated patterns (**line 11**). The pattern that results in the highest $E_S$ is designated as the best kernel (*bestfit_pattern*) for the root layer (**lines 11-14**). This process is repeated for all *k×k* kernels in the root layer of the groups.

**ALGORITHM 4: K×K KERNEL COMPRESSION**

    **Inputs:** *k×k Kernel Weights* ($K_W$)
    **Outputs:** *compressed k×k kernels, bestfit_pattern*
1  **best_Es** $\leftarrow \emptyset$
2  **bestfit_kernel** $\leftarrow \emptyset$
    *quant_bit* $\leftarrow$ array of quantization range
3  *temp_kernel* $\leftarrow$ *create a zeros array of shape* $K_W$
4  *positions [] $\leftarrow$ pattern_generator(n, $K_W$.shape[-1])*
5  **for** *p in positions:*
6    *temp_kernel [p] = temp_array [p]*
7    *temp_kernel [p] = $K_W$ [p]*
8    **for** *q in quant_bit:*
9      *compressed_kernel, sqnr =mp_quantizer (temp_kernel, q)*
10     *$K_W$ = compressed_kernel*
11     $E_S \leftarrow$ *calculate_ES(Mc, sqnr)*
12     **if** $E_S$ > *best_$E_S$*:
13       *best_$E_S$ = $E_S$*
14       *bestfit_kernel [$l_r$]= $K_W$*
15    **end**
16  **end**

**ALGORITHM 5: 1×1 KERNEL COMPRESSION**

    **Inputs:** *1×1 Kernel Weights* ($K_W$)
    **Outputs:** *compressed 1×1 kernel, bestfit_pattern*
1  *fl = [w for w in row for row in $K_W$] #flatten list of 1×1 kernel weights*
2  **best_Es** $\leftarrow \emptyset$
3  *temp_array* $\leftarrow \emptyset$
4  *bestfit_kernel* $\leftarrow \emptyset$
5  *quant_bit* $\leftarrow$ array of quantization ranges
6  *temp_kernel* $\leftarrow$ *create a zeros array of shape* $K_W$
7  **for** *i in range (0, len (fl), k):*
8    *t1 = fl [i : i+k]*
9    **if** *t1.shape[0] == k* **then**
10     *t1 = t1.reshape(k,k)*
11     *temp_array.append(t1)*
12    **else**: *temp_array.append(t1=0);*
13  **end**
14  *positions [] $\leftarrow$ pattern_generator(n, $K_W$.shape[-1])*
15  **for** *p in positions:*
16    *temp_kernel [p] = temp_array [p]*
17    **for** *q in quant_bit*
18      *compressed_kernel, sqnr = mp_quantizer(temp_kernel, q)*
19      *$K_W$ = Flatten(compressed_kernel)*
20      $E_S \leftarrow$ *calculate_$E_S$ ($M_c$, sqnr)*
21      **if** $E_S$ > *best_$E_S$*:
22       *best_$E_S$ = $E_S$*
23       *bestfit_kernel [$l_r$]= $K_W$*
24    **end**
25  **end**

For *1×1* kernels, even though these are abundant in most modern deep neural networks, their compression is often overlooked. To compress these kernels, we use a *1×1* to *k×k* transformation algorithm that enables grouped pruning and quantization. Algorithm 5 presents the pseudocode for this transformation and compression process. First, we reshape the *1×1* kernel ($K_W$) from the root layer, as described in Algorithm 1, flattening them and storing the result in *flatten_list* (**line 1**). We then iterate through *flatten_list*, grouping the values into sets of *k* weights, which are reshaped into a *k×k* weight matrix and stored in *temp_array* (**lines 7-13**). Next, we use our *pattern_generator* to randomly generate a pattern as per the number of non-zero elements and the resized kernel size (Algorithm 2; **line 14**). Utilizing the positions (*rows, cols*) of non-zero kernels identified from the generated patterns, we compress and quantize the *temp_kernels* in *temp_array* (**lines 15-25**). For quantizing *temp_kernel*, we use the *mp_quantizer* (Algorithm 6), which considers various bit-size alternatives (**line 18**). The modified kernel is then applied back to the kernel weights ($K_w$) by flattening the weights back to a *1×1* format (**line 19**). Afterwards, we calculate the efficiency score $E_s$ (eq. (2); discussed later), for all the generated kernels (**line 20**). The kernel with the highest $E_S$ is selected as the best kernel (*bestfit_pattern*) for the root layer (**lines 21-23**).

For both *1×1* and *k×k* kernels, given that the layers in the root group have coupled channels, the *bestfit_pattern* is employed to apply quantization to the leaf layers within the root group. The process of pruning, quantization, and kernel selection creates a large search space due to the different quantization bits and kernel combinations. To reduce the computational complexity of this exploration, we compress only the group root layers from *group_init*, which significantly limits the number of potential combinations to explore. By focusing on the root layers, we can more efficiently prune and quantize the kernels, reducing the computational burden of testing all layers and configurations. After applying the best pattern to the root layers, these optimized kernels are also applied to the subsequent leaf layers, making the algorithm more computationally feasible while still achieving effective compression and quantization.

*2) Mixed Precision Quantization*

We perform symmetric quantization, where we map the floating-point representation of each kernel weight to an equivalent integer space, such that both the real (floating point) and integer space is centered around 0. This approach enhances memory efficiency, especially in pruned models, by treating both positive and negative values equally, thus speeding up inference. Symmetric quantization can also result in higher Signal-to-Quantization-Noise Ratio (SQNR), preserving accuracy while minimizing quantization error. Symmetric quantization enables faster, more power-efficient inference by leveraging fixed-point arithmetic, improving throughput and latency in embedded platforms [35].

---

**ALGORITHM 6: *MP_QUANTIZER***
*Inputs: temp_kernel*
*Outputs: compressed_kernel, sqnr*
1   $x$ = temp_kernel
2   $α_x$ = max(|min(x)|,|max(x)|)
3   max_value = $2^{(quant\_bit - 1)} - 1$ # quant_bit is the integer bit quantization
4   min_value = $-(2^{(quant\_bit - 1)} - 1)$
5   scale = $α_x$ /max_value
6   $x_q$ = round(x/scale)
7   $x_q$ = clip($x_q$, min_val, max_val)
8   sqnr = var(x)/var(x - $x_q$)

---

Algorithm 6 presents the pseudocode for our quantization algorithm (*mp_quantizer*). We begin by copying the pruned *temp_kernel* to *x* and then calculating the scaling factor ($α_x$) for the kernel, which enables the mapping of continuous data to discrete representations while maintaining the integrity of the original information (**lines 1-2**). The scaling factor is the maximum absolute value of either the minimum or maximum of *x*, ensuring that the quantization scale will be appropriate for the range of values in the kernel. We calculate the maximum quantization value (*max_value*) and the minimum quantization value (*min_value*) based on the desired number of quantization bits (*quant_bit*) (**line 3-4**). We then compute the *scale* as the ratio of $α_x$ to the *max_value*, which determines the factor by which the kernel values will be scaled during quantization (**line 5**). The scale is then used to compute the quantized weight ($x_q$) by dividing the original values by the scale and rounding them to the nearest integer (**line 6**). We then apply a clipping operation to ensure that the quantized values remain within the range of (*min_value, max_value*) (**line 7**). Finally, the error between the pruned weight (*x*) and the quantized weight ($x_q$) is used to compute the SQNR, which provides a measure of the quantization error relative to the original kernel (**line 8**). The final outputs of the algorithm are the quantized kernel weights and the SQNR.

Our kernel compression algorithms (Algorithms 4 and 5) employ the *mp_quantizer* (Algorithm 6) along with iterative search through the *quant_bit* array to find the best $E_s$, to implement a mixed-precision quantizer. Through this approach, we can allow lower precision (e.g., 4 bits) for less significant kernels and higher precision (e.g., 16 bits) for more important kernels, to balance model footprint and accuracy.

*3) Efficiency Score*

We compute the efficiency score ($E_S$) of the model after updating the compressed kernel weight to the model $M_c$ as:

$$E_s = α.sqnr + β.\left(\frac{1}{Latency}\right) + γ.\left(\frac{1}{Energy}\right) \quad (2)$$

where *α, β, γ* are weights between [0,1] that determine the importance of each component in the efficiency score. We then calculate the on-device latency and energy of the model and use these values and calculate the $E_S$ of the model.

## V. EXPERIMENTAL RESULTS

In this section, we present results of prototyping and implementing our proposed *UPAQ* framework on the Jetson Orin Nano embedded platform and an Nvidia RTX 4080 workstation. We also contrast *UPAQ* with state-of-the-art compression techniques for 3D ODs.

*A. Experimental setup*

Our framework is evaluated on two state-of-the-art pretrained 3D ODs: 1) *PointPillars*, which uses pointcloud LiDAR data with 4.8 million parameters and an inference time of 35.98ms for the uncompressed model on Jetson Orin; and 2) *SMOKE*, which uses image-based input with 2D to 3D uplifting, consisting of 19.51 million parameters and 173 layers, with an inference time of 127.48ms for the uncompressed model on Jetson Orin. We implemented and tested the framework using PyTorch and TensorRT, on an Nvidia RTX 4080 workstation, and then deployed the model on the Jetson Orin. We calculate the power consumption of these models using NVpower tool [36]. The evaluation metrics include: 1) compression ratio; 2) mAP; 3) inference time; and 4) energy usage. We use the KITTI automotive dataset [33], split 80:10:10 for training, validation, and testing of both LiDAR pointcloud and RGB images.

**Table 2.** Comparison of *UPAQ* with base (uncompressed) model and state-of-the-art compression frameworks for the Pointpillars and SMOKE 3D ODs.

| Models | Metrics | | Base Model | Ps&Qs [20] | CLIP-Q [21] | R-TOSS [23] | LIDAR-PTQ [22] | UPAQ (LCK) | UPAQ (HCK) |
|---|---|---|---|---|---|---|---|---|---|
| **PointPillars** | Compression | | 1× | 1.89× | 1.84× | 4.07× | 3.25× | **4.92×** | **5.62×** |
| | mAP | | 78.96 | 83.67 | 79.68 | 85.26 | 78.90 | **86.15** | **84.25** |
| | Inference time (ms) | RTX 4080 | 5.72 | 5.17 | 5.26 | 5.69 | 4.25 | **2.37** | **1.70** |
| | | Jetson Orin | 35.98 | 32.061 | 35.07 | 35.94 | 29.65 | **19.96** | **18.23** |
| | Energy Usage (J) | RTX 4080 | 0.875 | 0.658 | 0.716 | 0.871 | 0.567 | **0.371** | **0.327** |
| | | Jetson Orin | 0.863 | 0.782 | 0.841 | 0.862 | 0.711 | **0.472** | **0.417** |
| **SMOKE** | Compression | | 1× | 1.95× | 1.84× | 4.25× | 3.57× | **4.23×** | **5.13×** |
| | mAP | | 29.85 | 31.03 | 30.45 | 32.56 | 30.23 | **36.65** | **35.49** |
| | Inference time (ms) | RTX 4080 | 28.36 | 23.72 | 25.48 | 24.98 | 12.75 | **9.67** | **8.23** |
| | | Jetson Orin | 127.48 | 93.65 | 87.28 | 98.87 | 86.27 | **71.35** | **68.45** |
| | Energy Usage (J) | RTX 4080 | 8.95 | 7.79 | 8.63 | 4.37 | 4.79 | **3.21** | **2.83** |
| | | Jetson Orin | 25.85 | 19.21 | 17.87 | 20.84 | 18.25 | **15.62** | **13.80** |

We evaluate two variants of the *UPAQ* framework: *1) UPAQ (HCK)*: which is biased towards higher compression, with fewer non-zero weights per kernel (e.g., 2 non-zero values for a 3×3 kernel) and more aggressive quantization with lower quantization bitwidths (e.g., a mix of 4 and 8 bits); and 2) *UPAQ (LCK)*: which is biased towards greater accuracy, with more non-zero weights than HCK (e.g., 3 non-zero values for a 3×3 kernel), and less aggressive quantization (e.g., a mix of 8 and 16 bits). The quantization bits (*quant_bits*) considered for experiments vary from 4 to 16 and we also set the weights in $E_s$ to be $\alpha=0.3, \beta=0.4, \gamma=0.3$ so that we give higher significance to minimizing model latency in our optimizations.

### B. Evaluation results of UPAQ compression framework:

We compared our *UPAQ* framework with the uncompressed Base Model (BM) and four state-of-the-art approaches. These include Ps&Qs (PQ) [21] which uses quantization-aware pruning with iterative pruning and pre-layer quantization using the same number of quantization bits. We also consider Clip-Q [22], which applies clipping, partitioning, and quantization. In this method, clipped weights are pruned, and non-clipped weights are quantized. We also consider R-TOSS [24] which applies entry-pattern based semi-structured pruning. Lastly, we consider Lidar-PTQ [23] that uses PTQ with max-min calibration and adaptive rounding for weight quantization in 3D ODs. Table 2 summarizes our evaluation results which will be discussed in more detail next.

In terms of accuracy, for PointPillars, *UPAQ* (LCK) achieves the best mAP of 86.15, surpassing Ps&Qs (83.67) and R-TOSS (85.26), while *UPAQ* (HCK) is close with an mAP of 84.25. For the SMOKE model, *UPAQ* (LCK) also delivers the highest mAP of 36.65.

In terms of compression (Table. 2), for the PointPillars model, *UPAQ* (HCK) achieves the highest compression ratio of 5.62×, outperforming Ps&Qs (1.89×), CLIP-Q (1.84×), LIDAR-PTQ (3.25×) and R-TOSS (4.07×). For the SMOKE model, *UPAQ* (HCK) also has the highest compression ratio of 5.13×, outperforming all other frameworks.

In terms of inference speedup on the Jetson Orin (Fig. 4), for PointPillars, *UPAQ* (HCK) achieves an inference time of 18.23ms, which is 1.97× faster than the base model (35.98ms) and faster than all other frameworks. *UPAQ* (LCK) reduces inference time to 19.96ms, which is 1.81× faster than the base model. For SMOKE, *UPAQ* (HCK) delivers an inference time of 68.45ms, which is 1.86× faster than the base model and faster than all other frameworks. *UPAQ* (LCK) also improves on the base model by 1.78× (an inference time of 71.35ms).

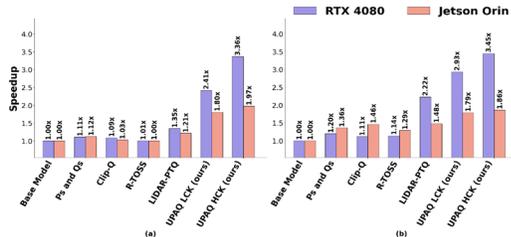

**Fig 4.** Comparison of speedups achieved in (a) PointPillars and (b) SMOKE models after using the compression frameworks.

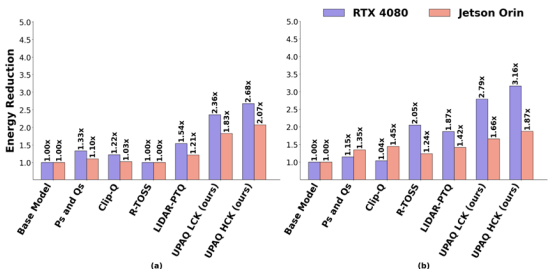

**Fig 5.** Comparison of reduction in energy usage achieved in (a) PointPillars and (b) SMOKE models after using the compression frameworks

In terms of energy usage on the Jetson Orin (Fig. 5), *UPAQ* (HCK) for PointPillars uses 0.417 J, which is 2.07× lower than the base model (0.863 J) and more efficient than Ps&Qs (0.782 J), CLIP-Q (0.841 J), LIDAR-PTQ (0.711 J) and R-TOSS (0.862 J). *UPAQ* (LCK) uses 0.472 J, 1.83× more efficient than the base model. For SMOKE, *UPAQ* (HCK) uses 15.62 J, which is 1.87× lower than the base (25.85 J) and more efficient than the other frameworks. *UPAQ* (LCK) reduces energy to 13.80 J, which is 1.66× more efficient than the base model.

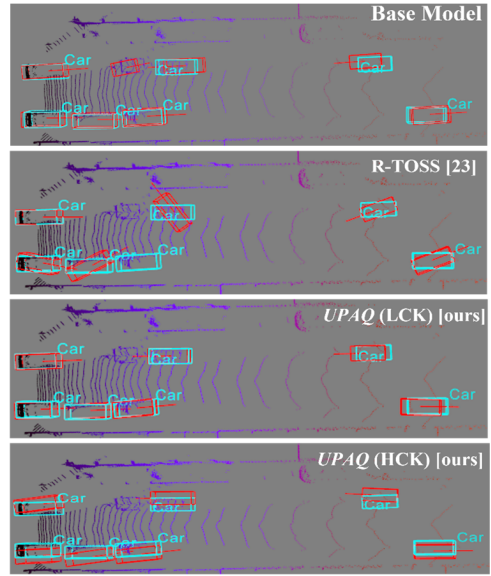

**Fig 6.** Comparison of output achieved in PointPillars model after using various compression frameworks. Blue bounding boxes indicate the ground truth positions of cars, while red boxes show each frameworks predictions.

Lastly, in Fig. 6, we compare object detection predictions across various frameworks using 3D pointcloud data from the KITTI dataset. In the figures, blue bounding boxes indicate the ground truth positions of cars, while red boxes show each frameworks predictions. We have selected three of the best performing models in terms of prediction accuracy, R-TOSS, *UPAQ* (HCK), and *UPAQ* (LCK), to be contrasted against the Base Model performance of PointPillars. For R-TOSS (85.26 mAP), even though it predicts all cars in the scene, it has misalignments in the results, which can lead to incorrect 3D positioning in driver assistance applications. *UPAQ* (LCK) achieves higher accuracy (86.15 mAP), with bounding boxes closely aligned with the ground truth and no extraneous predictions. *UPAQ* (HCK) also performs well, with minimal deviations from the true positions, closely matching *UPAQ* (LCK) in precision (84.25 mAP). Overall, our *UPAQ* frameworks, particularly *UPAQ* (LCK), demonstrate superior accuracy and reliability in 3D object detection, over other frameworks.

*In summary*, if accuracy is the priority, we recommend the LCK configuration for its superior mean Average Precision (mAP), despite slightly lower compression and speedup compared to HCK. For maximizing compression, inference speed, and energy efficiency, especially in resource-constrained environments, HCK is the better choice, offering higher compression, faster inference, and lower energy usage, at the cost of a slight reduction in accuracy.

## VI. CONCLUSIONS

In this paper, we introduce *UPAQ*, a novel 3D object detection (OD) compression framework that aims to preserve model accuracy while significantly reducing storage requirements and computational (performance, energy) overheads. Our comprehensive experimental results and comparative analyses demonstrate that *UPAQ* consistently outperforms state-of-the-art frameworks in terms of compression ratios, inference speed, and energy efficiency, while achieving a notable increase in mAP compared to baseline 3D OD models for LiDAR and camera data. The framework effectively minimizes computational costs during both compression and inference, facilitating a more efficient model compression process. Specifically, our experiments on the Jetson Orin platform show that *UPAQ* achieves model compression rates of 5.62× for the PointPillars 3D OD model and 5.13× for the SMOKE 3D OD model, while also surpassing the original model in both mAP and inference speed, underscoring the efficacy of our approach. Our ongoing work is looking at combining deep learning techniques for anomaly detection [37], [38] and sensor deployment [39], [40] with optimized object detection.